\documentclass{article}
\usepackage[utf8]{inputenc}
\usepackage[T1]{fontenc}
\usepackage[margin=1in]{geometry}
\usepackage{graphicx}
\usepackage{subfig}
\usepackage{amsmath, amssymb, amsthm}
\usepackage{url}
\usepackage{hyperref}
\usepackage{listings}
\usepackage{xcolor}
\usepackage{amsmath}
\usepackage{amssymb}
\usepackage{algorithm}
\usepackage{algpseudocode}
\usepackage{amsmath}
\usepackage{multirow}      
\usepackage{booktabs}      
\usepackage{array}         
\usepackage{etoolbox}  
\AtBeginEnvironment{algorithmic}{\catcode`\&=12\relax}
\DeclareUnicodeCharacter{2212}{-}
\newcommand{\First}{M. Parsa Dini \\ \texttt{mp.dini@ee.sharif.edu}}
\newcommand{\Second}{Human Jafari \\ \texttt{houman.jafari627@sharif.edu}}

\graphicspath{{figures/}}
\title{
    \textbf{ADAPTIVE-$\lambda$ SISS in MACHINE UNLEARNING for DDPM \& VAE} \\[0.5em]
}
\author{\First\and\Second
}
\date{} 
\begin{document}
\maketitle

\begin{abstract}
\textbf{Machine Unlearning (MU)} is critical for large generative models, such as VAEs
and DDPMs,
to enforce the \textit{ right to be forgotten} and ensure safe content generation without expensive retraining. Existing methods, like the Static-$\lambda$ SISS framework
for Diffusion Models, rely on a fixed data mixing parameter $\lambda$, which is suboptimal as the necessary unlearning rate varies dynamically across samples and training stages. 

We introduce Adaltive-$\lambda$ SISS, a novel approach that transforms the mixing parameter $\lambda$ from a fixed constant into a latent variable whose optimal value is dynamically inferred at each unlearning step. We leverage a Variational Inference-based objective
, employing a small neural network to learn an adaptive posterior distribution $q_\phi(\lambda)$. This network conditions on context features derived from the SISS loss components, enabling simultaneous and robust optimization of the diffusion model and the $\lambda$-inference mechanism.


Furthermore, we propose novel extensions to this unlearning paradigm. First, we explore a unified unlearning objective that leverages the Score Forgetting Distillation (SFD) framework’s data-free efficiency and the SISS framework’s direct gradient control, creating a hybrid approach that exploits the strengths of both. Second, to enhance the robustness and fine-grained control of the unlearning process, we introduce a Reinforcement Learning (RL) framework that learns an optimal, sequential unlearning policy, treating the unlearning operation as an action in a state space defined by the model's current memory of the forget-set.
\end{abstract}

\section{Introduction}


 Modern generative models, ranging from VAEs and DDPMs to large-scale systems such as Stable Diffusion are trained on massive, high-dimensional datasets that frequently contain sensitive personal information. As a result, growing regulatory and ethical pressures have intensified the need for machine unlearning: the ability to remove or suppress the influence of specific data points, concepts, or classes after training.
 
 Practical motivations for unlearning span several dimensions. Individuals may demand that facial images resembling them in datasets such as CelebA be removed from the learned model. More broadly, erasing semantic concepts or class labels enables a pathway toward controllable editing \cite{orgad2024editing} and concept-level modification within generative models. In addition, unlearning supports safe guidance, where a model avoids sampling from inappropriate or undesirable regions of the data distribution. These considerations, together with the infeasibility of retraining large models from scratch, have made unlearning an increasingly active research direction.
 
 Large Language Models \& Large generative models, including Diffusion Models (DDPMs, VAEs, Stable Diffusion), are typically trained on massive, high-dimensional datasets. Retraining these models from scratch is computationally prohibitive, necessitating the development of efficient and effective unlearning methodologies. Furthermore, unlearning serves as a foundation for concept erasure and general data editing, while also bolstering defense against Membership Inference Attacks (MIA).

Our work significantly advances machine unlearning in generative models through several key innovations. Beyond the adaptive SISS framework, we explore methods that bridge existing limitations. Specifically, we investigate a hybrid objective that merges the robust, data-dependent unlearning control of the SISS framework with the efficient, conditional score-alignment of SFD. This synthesis aims to achieve rapid and effective concept erasure while maintaining high generative quality. Building on this, we introduce a Reinforcement Learning-based machine unlearning method. This RL approach allows the unlearning procedure itself to be optimized as a sequential decision-making process, enabling the model to learn context-aware, temporally optimal updates that maximize forgetting while minimizing performance degradation on the retain set. This is a crucial step toward creating truly robust and efficient unlearning protocols.

\section{Related Works}
Several approaches to machine unlearning have been explored. A straightforward strategy applies gradient ascent\cite{thudi2022unrolling} on a designated forget-set to counteract learned influence. Other methods operate in latent space, steering samples toward alternative semantic directions. Weight-saliency-based approaches attempt to identify and suppress parameter subsets most responsible for memorization. More recently, works such as Ermon et al. (SISS framework) propose relaxing the goal of exact unlearning by instead attenuating the contribution of forgotten data in a controlled manner, acknowledging that strict removal often harms generative fidelity.

However, existing frameworks exhibit notable limitations. Many assume that forgotten samples should map to pure noise like \cite{li2024machine_unlearn}, which leads to semantically broken or incoherent generations, contradicting the practical goal of replacing forgotten content with meaningful high-probability alternatives.

In our work, we focus on several key ideas. First, we address the poor forgetting and sample-generation performance of the SISS framework by introducing additional components to its design. In Ermon et al.’s work, the parameter 
$\lambda$; the Bernoulli variable controlling the probability of sampling from the forget-set during the DDPM backward process was fixed at $\lambda=\frac{1}{2}$. When $\lambda$ approaches $0$ or $1$, SISS effectively draws its noisy latents almost entirely from the retain set or from the forget set, respectively, producing latents that are unlikely to be sampled from the opposite distribution. Therefore, this fixed choice can be problematic, as different images may require different values of $\lambda$. To remedy this, we propose sampling $\lambda$ from a posterior distribution, which leads to significantly improved performance compared to the original method.

Second, we revisit the label-assignment strategy used for guidance within the SFD framework, proposing improvements for more effective unlearning.

Third, we explore a unified approach that combines ideas from both the first and second directions.

Finally, we introduce a reinforcement-learning–based machine-unlearning method to further enhance unlearning efficiency and robustness.
\section{Problem Formulation}
Let $\mathcal{D} = \{ x_i \}_{i=1}^{n}$ denote the original training data. Suppose we wish to unlearn a subset $\mathcal{F} = \{ a_j \}_{j=1}^{k} \subset \mathcal{D}$. Retraining the full model, parameterized by $\theta$, on the retain-set $\mathcal{R} = \mathcal{D} \setminus \mathcal{F}$ is typically infeasible due to computational cost and model scale. The ideal exact machine unlearning objective is to achieve parity between the model trained on the full dataset and a hypothetical model trained only on the retain-set:
$$
p_{\theta}(x \mid \mathcal{D}) \approx p(x \mid \mathcal{R}).
$$
To approximate this goal efficiently, most unlearning methods adopt a bi-criteria optimization based on the empirical risk minimization framework, balancing the effort to suppress the forget-set's influence while preserving performance on the retain-set:
$$
\theta^* = \arg\min_\theta \left\{ \mathcal{L}_{\text{retain}} - \alpha \mathcal{L}_{\text{forget}} \right\}.
$$
Here, $\mathcal{L}_{\text{retain}}$ and $\mathcal{L}_{\text{forget}}$ are loss functions or divergence measures defined over samples from the retain-set $\mathcal{R}$ and forget-set $\mathcal{F}$, respectively, and $\alpha > 0$ is a scalar hyperparameter controlling the forgetting strength. This general structure is the foundation for numerous state-of-the-art unlearning techniques. For instance, the SISS framework explicitly uses this adversarial loss structure on a mixture distribution, while the Score Forgetting Distillation (SFD) framework employs a variant where $\mathcal{L}_{\text{retain}}$ is a knowledge distillation loss and $\mathcal{L}_{\text{forget}}$ aligns the forget-concept score with a retain-concept score. Similarly, other approaches, such as those minimizing the KL divergence between the unlearned and vanilla model outputs on the retain-set and maximizing it on the forget-set, adhere to this core objective.

\section{Preliminaries}
In this section, first we will review VAE \& DDPM models and afterwards, we will continue with SISS \& SFD frameworks
\subsection{Variational Autoencoders (VAEs)}
Variational Autoencoders (VAEs) \cite{kingma2013auto, rezende2014stochastic} are a generative model that learns a latent variable representation of the data. They consist of two main components: an Encoder (or inference model), $q_{\phi}(\mathbf{z}|\mathbf{x})$, and a Decoder (or generative model), $p_{\theta}(\mathbf{x}|\mathbf{z})$.

The Encoder maps an input data point $\mathbf{x}$ to a distribution over the latent space $\mathbf{z}$, introducing the necessary stochasticity. The Decoder reconstructs the data $\mathbf{x}$ from a sample $\mathbf{z}$ drawn from this latent distribution. The model is trained to maximize the evidence lower bound (ELBO), which is a tractable lower bound on the log-likelihood of the data, $\log p(\mathbf{x})$.

The ELBO objective function is given by:
\begin{equation}
\mathcal{L}_{\text{VAE}}(\theta, \phi; \mathbf{x}) = \mathbb{E}_{q_{\phi}(\mathbf{z}|\mathbf{x})}\Big[\log p_{\theta}(\mathbf{x}|\mathbf{z})\Big] - D_{\text{KL}}\Big(q_{\phi}(\mathbf{z}|\mathbf{x}) \parallel p(\mathbf{z})\Big)
\end{equation}

\noindent The first term is the reconstruction loss, ensuring the decoder accurately recreates the input. The second term is the Kullback-Leibler (KL) divergence between the learned posterior $q_{\phi}(\mathbf{z}|\mathbf{x})$ and a prior distribution $p(\mathbf{z})$ (typically Gaussian), which acts as a regularizer to structure the latent space.
\subsection{Denoising Diffusion Probabilistic Models (DDPMs)}
\subsubsection{DDPM (VAE-Based Approach)}
The Forward Process is a fixed Markov chain that gradually adds Gaussian noise to the data $\mathbf{x}_0$ over $T$ time steps.
The transition probability is:
\begin{equation}
q(\mathbf{x}_t|\mathbf{x}_{t-1}) = \mathcal{N}\Big(\mathbf{x}_t; \sqrt{1-\beta_t}\mathbf{x}_{t-1}, \beta_t \mathbf{I}\Big)
\end{equation}

where $\beta_t$ is a predefined noise schedule. If this process continues for many steps ($T \to \infty$), the data distribution $p(\mathbf{x}_0)$ is diffused into a simple, known Gaussian distribution, $p(\mathbf{x}_T) \approx \mathcal{N}(\mathbf{0}, \mathbf{I})$.

The Reverse Process is a learned Markov chain that reverses the diffusion, gradually denoising the data from $\mathbf{x}_T$ back to $\mathbf{x}_0$. The transition probabilities are parameterized by a neural network (e.g., a U-Net):
\begin{equation}
p_{\theta}(\mathbf{x}_{t-1}|\mathbf{x}_t) = \mathcal{N}\Big(\mathbf{x}_{t-1}; \mu_{\theta}(\mathbf{x}_t, t), \Sigma_{\theta}(\mathbf{x}_t, t)\Big)
\end{equation}

In practice, the simplified training objective minimizes the mean squared error between the true noise $\mathbf{\epsilon}$ and the noise predicted by the model $\mathbf{\epsilon}_{\theta}$ \cite{ho2020denoising}:
\begin{equation}
\mathcal{L}_{\text{DDPM}}(\theta) =
\mathbb{E}_{t, \mathbf{x}_0, \boldsymbol{\epsilon}}
\left[
\left\|
\boldsymbol{\epsilon}
-
\boldsymbol{\epsilon}_{\theta}
\big(
\sqrt{\bar{\alpha}_t}\,\mathbf{x}_0
+
\sqrt{1-\bar{\alpha}_t}\,\boldsymbol{\epsilon},
t
\big)
\right\|_2^2
\right]
\end{equation}

\subsubsection{Score-Based DDPM \& SDE Approach}
This perspective focuses on learning the score function of the noisy data distribution, $s(\mathbf{x}, t) = \nabla_{\mathbf{x}}\log p_t(\mathbf{x})$. The score function is a vector field that, at any point $\mathbf{x}$, points towards the nearest high-probability region of the data distribution. The generative model $s_{\theta}(\mathbf{x}, t)$ is trained to approximate this true score function \cite{song2019generative}.

The training is done using the denoising score matching loss:
\begin{equation}
\mathcal{L}_{\text{SGM}}(\theta) =
\mathbb{E}_{t, \mathbf{x}_0, \mathbf{x}_t}
\left[
\left\|
\mathbf{s}_{\theta}(\mathbf{x}_t, t) 
- \nabla_{\mathbf{x}_t} \log q(\mathbf{x}_t \mid \mathbf{x}_0)
\right\|_2^2
\right]
\end{equation}

This framework unifies DDPMs by defining the noise injection and generation process via Stochastic Differential Equations (SDEs) \cite{song2020scorebased}.

The continuous Forward SDE for noise injection is:
$$
d\mathbf{x} = f(\mathbf{x}, t) dt + g(t) d\mathbf{w}
$$
where $f(\mathbf{x}, t)$ is the drift, $g(t)$ is the diffusion coefficient, and $\mathbf{w}$ is the standard Wiener process.

The corresponding Reverse SDE for sampling and generation is:
\begin{equation}
\mathrm{d}\mathbf{x}_t = \Bigl[ f(\mathbf{x}_t, t) - g(t)^2 \nabla_{\mathbf{x}_t} \log p_t(\mathbf{x}_t) \Bigr] \mathrm{d}t + g(t) \, \mathrm{d}\bar{\mathbf{w}}_t
\end{equation}
where $d\mathbf{\bar{w}}$ is the infinitesimal reverse-time Wiener process. By substituting the learned score $s_{\theta}(\mathbf{x}, t)$ for $\nabla_{\mathbf{x}}\log p_t(\mathbf{x})$ and solving the SDE numerically, new data samples are generated.
\subsection{SISS Framework}
The SISS framework utilizes a mixture distribution $q_\lambda(m_t|x_r, x_f)$ from which the noisy sample $m_t$ is drawn at time step $t$ (assume that $x_r$ is from retain-set and $x_f$ is from forget-set)
\begin{equation}
\forall \lambda \in [0,1] : q_\lambda(m_t|x_r,x_f) = (1-\lambda) q(m_t|x_r) + \lambda q(m_t|x_f)  
\end{equation}
where $q(m_t|x) = \mathcal{N}(m_t|\gamma_t x , \sigma_t^2 I)$.
It is like tossing a coin and if it was heads which happens with probability of $\lambda$, we sample from $\mathcal{F}$, otherwise, we sample $m_t$ from $\mathcal{R}=\mathcal{D} \backslash  \mathcal{F}$. 

The unlearning objective $\mathcal{L}_{\text{siss}}$ is defined using the retain and forget loss components, $\mathcal{L}_{\text{retain}}$ and $\mathcal{L}_{\text{forget}}$, which employ Importance Sampling (IS) to correct for sampling from $q_\lambda$:

\begin{equation}
\begin{aligned}
\mathcal{L}_{\text{retain}}(x_r, x_f, m_t, t; \theta) &= \frac{n}{n-k} \frac{q(m_t|x)}{q_\lambda(m_t|x_r,x_f)} \left\| \frac{m_t - \gamma_t x_r}{\sigma_t} - \epsilon_\theta(m_t,t) \right\|_2^2 \\
\mathcal{L}_{\text{forget}}(x_r, x_f, m_t, t; \theta) &= \frac{k}{n-k} \frac{q(m_t|a)}{q_\lambda(m_t|x_r,x_f)} \left\| \frac{m_t - \gamma_t x_f}{\sigma_t} - \epsilon_\theta(m_t,t) \right\|_2^2
\end{aligned}
\end{equation}

where $\epsilon_\theta(m_t,t)$ is the predicted noise by the UNet with parameters $\theta$. The $\frac{n}{n-k}$ and $\frac{k}{n-k}$ terms are dataset-size-related weighting terms, where $n$ is the total dataset size and $k$ is the size of the forget set.The overall unlearning loss is:\\

\begin{equation}
\begin{aligned}
\mathcal{L}_{\text{siss}} &= 
\mathbb{E}_{x_r \sim P_R} \, \mathbb{E}_{x_f \sim P_F} \, \mathbb{E}_{t \sim P_T} \, \mathbb{E}_{m_t \sim q_\lambda} 
\Big[
\mathcal{L}_{\text{retain}} - (1+s)\, \mathcal{L}_{\text{forget}}
\Big] \\
&=
\mathbb{E}_{x_r \sim P_R} \, \mathbb{E}_{x_f \sim P_F} \, \mathbb{E}_{t \sim P_T} \, \mathbb{E}_{m_t \sim q_\lambda} 
\Bigg[
\frac{n}{n-k} \cdot \frac{q(m_t \mid x_r)}{q_\lambda(m_t \mid x_r,x_f)} 
\Big\| \frac{m_t - \gamma_t x_r}{\sigma_t} - \epsilon_\theta(m_t,t) \Big\|_2^2 \\
&\quad - (1+s) \cdot \frac{k}{n-k} \cdot \frac{q(m_t \mid x_f)}{q_\lambda(m_t \mid x_r,x_f)} 
\Big\| \frac{m_t - \gamma_t x_f}{\sigma_t} - \epsilon_\theta(m_t,t) \Big\|_2^2
\Bigg]
\end{aligned}
\end{equation}

, where $s \ge 0$ is a scale factor for the forget loss. Here is the algorithm of SISS Framework:

\begin{figure}[h!]
\centering
\fbox{%
\begin{minipage}{0.95\columnwidth}
    \centering
    \vspace{0.5em}
    \textbf{Algorithm 1: Static-$\lambda$ SISS Unlearning for DDPMs}
    \vspace{0.5em}
    \hrule
    \vspace{1em}
    \begin{algorithmic}[1]
    \Require Dataset $\mathcal{D}$, Forget set $\mathcal{F}$, Retain set $\mathcal{R}$, Model $\varepsilon_\theta$, Noise $q$, mixture weight $\lambda$, learning rate $\eta$, scaling $s$, epochs $E$
    \For{$e = 1$ to $E$}
        \State Sample $x_r \sim \mathcal{R}$, $x_f \sim \mathcal{F}$, $t \sim P_T$
        \State Sample latent: $m_t \sim q_\lambda(m_t \mid x_r,x_f) = (1-\lambda)q(m_t|x_r) + \lambda q(m_t\mid x_f)$
        \State \textbf{Retain Loss:}
        \Statex \hspace{\algorithmicindent}
        $\mathcal{L}_{\mathrm{retain}} \gets \frac{|\mathcal{D}|}{|\mathcal{R}|} \frac{q(m_t|x_r)}{q_\lambda(m_t|x_r,x_f)}
        \left\| \frac{m_t - \gamma_t x_r}{\sigma_t} - \varepsilon_\theta(m_t,t) \right\|^2$
        \State \textbf{Forget Loss:}
        \Statex \hspace{\algorithmicindent}
        $\mathcal{L}_{\mathrm{forget}} \gets \frac{|\mathcal{D}|}{|\mathcal{F}|} \frac{q(m_t|x_f)}{q_\lambda(m_t|x_r, x_f)}
        \left\| \frac{m_t - \gamma_t x_f}{\sigma_t} - \varepsilon_\theta(m_t,t) \right\|^2$
        \State \textbf{Total loss:} $\mathcal{L}_{\mathrm{SISS}} \gets \mathcal{L}_{\mathrm{retain}} - (1+s)\mathcal{L}_{\mathrm{forget}}$
        \State Optionally clip/scale gradient of $\mathcal{L}_{\mathrm{forget}}$
        \State Update: $\theta \gets \theta - \eta \nabla_\theta \mathcal{L}_{\mathrm{SISS}}$
    \EndFor
    \end{algorithmic}
\end{minipage}%
}
\caption{Static-$\lambda$ SISS unlearning for DDPMs.}
\label{alg:static_siss}
\end{figure}
\subsection{SFD Framework}

The SFD (Score Forgetting Distillation) framework enables data-free machine unlearning in diffusion models by distilling knowledge from a pretrained model into a one-step generator while aligning scores of forget classes/concepts with retain ones. For a forget condition $c_f$ and override condition $c_o$ from the retain set $C_r$, the framework optimizes a generator $g_\theta(n, c)$ with $n \sim \mathcal{N}(0, I)$ such that the induced distribution $\mathcal{D}_{\theta, c_f}$ matches $p_{\text{data}}(x \mid c_o)$, while preserving $\mathcal{D}_{\theta, c_r} \approx p_{\text{data}}(x \mid c_r)$ for $c_r \in C_r$.

Noisy samples are generated via forward diffusion: $z_t = a_t x + \sigma_t \epsilon_t$ with $\epsilon_t \sim \mathcal{N}(0, I)$, $x \sim D_{\theta, c}$, where $a_t, \sigma_t$ follow the diffusion schedule. The pretrained model provides the score $s_\phi(z_t, c, t) = \nabla_{z_t} \ln p_{\text{data}}(z_t \mid c)$ and mean predictor $x_\phi(z_t, c, t)$. The generator's optimal score is $s_{\psi^*(\theta)}(z_t, c, t) = a_t x_{\psi^*(\theta)}(z_t, c, t) - z_t / \sigma_t^2$, where $\psi^*(\theta)$ minimizes the denoising score matching loss over fake data.

The core SFD loss aligns pretrained and generator scores across conditions:
\begin{equation}
\mathcal{L}_{\text{sfd}}(\theta; \phi, c_1, c_2) 
= \mathbb{E}_{z_t, t, x \sim D_{\theta, c_2}} \Bigl[ 
    \omega_t \Bigl\| s_\phi(z_t, c_1, t) - s_{\psi^*(\theta)}(z_t, c_2, t) \Bigr\|_2^2 
\Bigr],
\end{equation}
with re-weighting $\omega_t > 0$. This decomposes into distillation $\mathcal{L}_{\text{sfd}}(\theta; \phi, c_r, c_r)$ and forgetting $\mathcal{L}_{\text{sfd}}(\theta; \phi, c_o, c_f)$ components (which is quite the forget \& retain approach).

The overall objective balances retention and forgetting:
\begin{equation}
\mathcal{L}_{\text{sfd}} = \mathbb{E}_{c_r \sim C_r} \Bigl[
  \mathcal{L}_{\text{sfd}}(\theta; \phi, c_r, c_r) 
  + \lambda \, \mathcal{L}_{\text{sfd}}(\theta; \phi, c_o, c_f)
\Bigr],
\end{equation}
where $\lambda > 0$ are balancing hyperparameters. In practice, alternating optimization between $\theta$ and an approximate score network $\psi$ uses an equivalent form via Tweedie's formula:
\begin{equation}
\begin{aligned}
\hat{\mathcal{L}}_{\text{sfd}}\bigl(\theta, \psi; \phi, c_1, c_2, \alpha\bigr) 
&= (1 - \alpha) \omega_t a_t^2 \sigma_t^4 \left\| x_\phi(z_t, c_1, t) - x_\psi(z_t, c_2, t) \right\|_2^2 \\
&\quad + \omega_t a_t^2 \sigma_t^4 \left( x_\phi(z_t, c_1, t) - x_\psi(z_t, c_2, t) \right)^\top \! 
         \left( x_\psi(z_t, c_2, t) - x \right)
\end{aligned}
\end{equation}
with $\alpha \in \{1, 1.2\}$ and $\omega_t = \sigma_t^4 a_t^2 C \| x_\phi(z_t, t, c) - x \|_1^{\text{sg}}$ for stop-gradient. This data-free approach requires only the pretrained score network, enabling efficient unlearning without real or fully denoised samples.
\section{Our Contribution }
\subsection{Reforming SISS Framework}
The primary limitation of SISS is the fixed $\lambda$. We introduce a novel approach to learn an adaptive $\lambda$ at each unlearning step, making the process dynamic and sensitive to the current degree of forgetting and retention.

The SISS framework utilizes a weighted sampling distribution $q_\lambda(m_t|x,a)$ where $x$ is a retain-set sample and $a$ is a forget-set sample. The parameter $\lambda \in [0, 1]$ controls the mixture. Our key innovation is to replace the fixed $\lambda$ with a neural network that learns its optimal value during the unlearning process. This learned $\lambda$ is designed to be adaptive to the current state of the unlearning process, encoded by context features derived from the SISS loss components.
\subsubsection{Variational Inference for $\lambda$}
We treat $\lambda$ as a latent variable and utilize a separate, small neural network, parametrized by $\phi$, to infer its distribution $q_\phi(\lambda)$ given a set of features describing the current state of the unlearning process. This leads to a Variational Autoencoder (VAE)-like objective applied to the SISS loss:

\begin{equation}
\begin{aligned}
\mathcal{L}_{\text{ELBO}} &= \mathbb{E}_{\lambda \sim q_\phi(\lambda)} \Big[ \mathcal{L}_{\text{siss}}(\lambda) \Big] + \beta D_{\text{KL}} \Big(q_\phi(\lambda) || p(\lambda) \Big)  \\
&= \mathbb{E}_{x_r \sim P_R} \mathbb{E}_{x_f \sim P_F} \mathbb{E}_{t \sim P_T} \mathbb{E}_{\lambda \sim q_\phi(\lambda)} \mathbb{E}_{m_t \sim q_\lambda} 
\Bigg[ \\
&\quad \frac{n}{n-k} \cdot \frac{q(m_t|x_r)}{q_\lambda(m_t|x_r,x_f)} \left\| \frac{m_t - \gamma_t x_r}{\sigma_t} - \epsilon_\theta(m_t,t) \right\|_2^2 \\
&\quad - (1+s) \cdot \frac{k}{n-k} \cdot \frac{q(m_t|x_f)}{q_\lambda(m_t|x_r,x_f)} \left\| \frac{m_t - \gamma_t x_f}{\sigma_t} - \epsilon_\theta(m_t,t) \right\|_2^2 \Bigg] + \beta D_{\text{KL}} \Big(q_\phi(\lambda) || p(\lambda) \Big)
\end{aligned}
\end{equation}

his loss simultaneously optimizes the diffusion model parameters $\theta$ through $\mathcal{L}_{\text{siss}}(\lambda)$ and the inference network parameters $\phi$ through the Kullback-Leibler (KL) divergence term, $D_{\text{KL}}$, which regularizes the posterior $q_\phi(\lambda)$ towards a prior $p(\lambda)$. We can assume $p(\lambda)$ is a simple uniform or Gaussian prior.
\subsubsection{ Context Feature Vector and Reparameterization}

\begin{figure}[h!]
    \centering
    \includegraphics[scale=0.45]{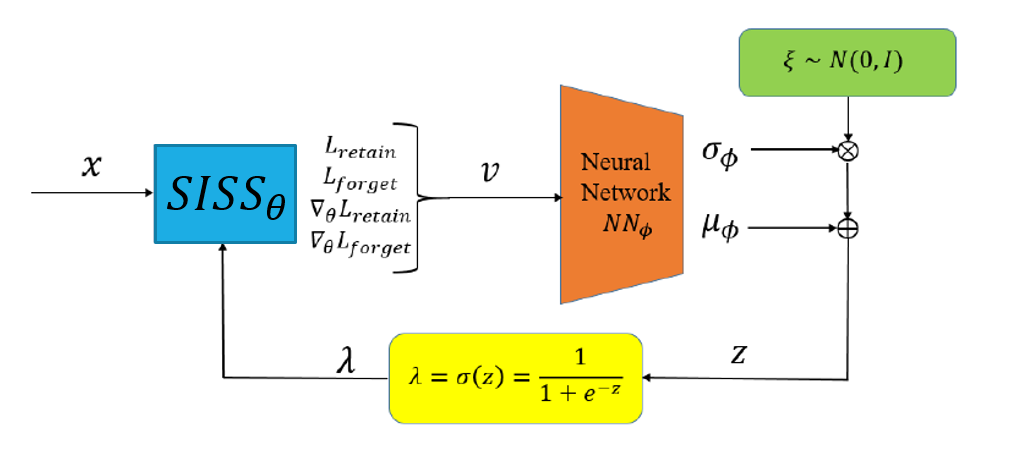} 
    \caption{In this image the process that how our model works is depited. the model uses the retain and forget losses of SISS and uses them to encode a latent variable $z$, which later by sampling from posterior distribution we get $z$ and by feeding it into a sigmoid function we get the $\lambda$ we wanted.  }
    \label{fig:our_arch}
\end{figure}

To make $\lambda$ adaptive, the inference network $q_\phi(\lambda)$ takes a context vector $v$ as input, which encapsulates the current unlearning feedback:
$$v = \Big[\mathcal{L}_{\text{retain}}, \mathcal{L}_{\text{forget}}, ||\nabla_\theta\mathcal{L}_{\text{retain}}||, ||\nabla_\theta\mathcal{L}_{\text{forget}}||\Big]
$$
The network outputs the parameters $(\mu_\phi(v), \sigma_\phi(v))$ for a Gaussian distribution on a latent variable $z$. We use the reparameterization trick to ensure that the gradient can flow back to $\phi$:$$z = \mu_\phi(v) + \xi \cdot \sigma_\phi(v) , \quad \xi \sim \mathcal{N}(0,1)$$The final adaptive $\lambda$ is obtained by passing $z$ through a sigmoid function to constrain it to the interval $[0, 1]$ (The architecture of our framework is depicted in Figure~\ref{fig:our_arch}): $$\lambda = \sigma(z) = \frac{1}{1+e^{-z}}$$.

This approach has two benefits, one clearly makes the SISS framework much mature and with better performance, secondly it does not use any other pretrained model and it is very easy to calculate because we made a gaussian prior assumption which makes the KL divergence a very easy differentiable function and also for generation of $\lambda$, we used sigmoid function, another easily  differentiable function.

For the context feature vector $v$, we selected the retain and forget losses along with the norms of their corresponding gradients. In practice, this combination provided a sufficiently informative and stable signal for the inference network to learn an effective mapping to the adaptive $\lambda$. Within this setup, the model can be trained end-to-end to infer an optimal balance between forgetting and retention based solely on these core optimization dynamics. Although this minimal set of features already yielded strong performance, an alternative extension would be to augment $v$ with semantic or representation-level features such as embeddings or concept-specific activations. Such additions could potentially enrich the contextual information available to the inference network, although we leave this direction for future exploration.

\subsubsection{ Backpropagation and Gradient Update}
The gradient of ELBO with respect to the inference network parameters $\phi$ is calculated using the chain rule, using the reparameterization trick:

\begin{equation}
\frac{\partial \mathcal{L}_{\text{ELBO}}}{\partial \phi} = \mathbb{E}_{\xi \sim \mathcal{N}(0,1)} \Bigg[
\frac{\partial \mathcal{L}_{\text{siss}}}{\partial \lambda} \cdot \frac{\partial \lambda}{\partial z} \Bigg(
\frac{\partial z}{\partial \mu}\cdot \frac{\partial \mu}{\partial \phi} +
\frac{\partial z}{\partial \sigma}\cdot \frac{\partial \sigma}{\partial \phi}
\Bigg) + \beta \frac{\partial D_{\text{KL}}}{\partial \phi}
\Bigg]
\end{equation}
The intermediate derivatives are:$$\frac{\partial \lambda}{\partial z} = \lambda(1-\lambda), \quad \frac{\partial z}{\partial \mu}=1, \quad \frac{\partial z}{\partial \sigma} = \xi
$$

We can also observe that our changes to the original SISS framework do not add significant computational cost. This is clear from the derivatives above and because the derivative of the KL divergence is easily computed.

\subsection{Adaptive-$\lambda$ Multi-Class Concept Erasure for Score-Based Unlearning}
While SISS is VAE-based, the adaptive $\lambda$ concept can enhance score-function-based unlearning methods like Score Forgetting Distillation (SFD). SFD, which seeks to approximate the unlearned score function $\nabla_x \log p(x)$ via distillation, is generally more effective for concept erasure (e.g., erasing a label $c_f$ in favor of $c_o$).The SFD objective combines a retaining distillation loss and a forgetting loss:$$\mathcal{L}_{\text{sfd}}(\theta, \psi) \propto \hat{\mathcal{L}}_{\text{sfd}}(\theta, \psi; \dots, c_r, c_r,\alpha ) + \lambda \hat{\mathcal{L}}_{\text{sfd}}(\theta, \psi; \dots, c_o, c_f,\alpha )$$where $\psi$ are parameters for a teacher network and $c_r$ is the retain class.We propose a multi-class extension using an adaptive $\lambda$ for the concept erasure portion:
$$\mathbb{E}_{\lambda \sim q_\phi(\lambda)} \Bigg[
\hat{\mathcal{L}}_{\text{sfd}}(\theta, \psi;\phi, c_r, c_r,\alpha ) +
\sum_{c_o \in \mathcal{C}_{\text{retain}}}
\frac{q(m_t|c_o)}{q_\lambda(x,\mathcal{C})}
\hat{\mathcal{L}}_{\text{sfd}}(\theta, \psi;\phi, c_o, c_f,\alpha )
\Bigg]$$
Here, the forgetting process of $c_f$ is directed towards a mixture of all retain classes $c_o \in \mathcal{C}_{\text{retain}}$, weighted by an adaptive $\lambda$. The importance weight $\frac{q(m_t|c_o)}{q_\lambda(x,\mathcal{C})}$ is adapted to the multi-class context, guiding the model to a high-probability manifold other than the one associated with the forget class.
\subsection{Unlearning as a Sequential Decision Process: Reinforcement Learning for $\lambda$ Selection}

\label{sec:mdp_unlearning}

The adaptive SISS formulation in Section~5.1 makes
$\lambda$ depend on the current training context, but the optimization is still
essentially myopic: at each step, $\lambda$ is chosen to reduce the
instantaneous SISS loss on the current mini-batch. In practice, however, a
particular choice of $\lambda$ can have long-term effects on the model.
Choosing an overly aggressive forgetting weight early in training can
irreversibly damage the retain distribution, while overly conservative values
may lead to under-forgetting even after many optimization steps.

To explicitly capture this long-term trade-off between forgetting and
retention, we model the unlearning process as a finite-horizon Markov Decision
Process (MDP), in which adaptively selecting $\lambda$ becomes a sequential
decision problem, and the environment dynamics are induced by the gradient
updates of the diffusion model.

\subsubsection{MDP formulation}

We define an MDP
$\mathcal{M} = (\mathcal{S}, \mathcal{A}, \mathcal{T}, r, \gamma)$ that
describes the evolution of the model during unlearning.

\paragraph{Episodes and time index.}
We consider unlearning as a finite-horizon process of $T$ optimization
steps. One \emph{episode} corresponds to a full unlearning run of $T$ gradient
updates. At each step $t = 1, \dots, T$, an agent observes the current
unlearning state of the model, chooses a forgetting weight $\lambda_t$, and
the model parameters are updated using the SISS loss evaluated with that
choice of $\lambda_t$.

\paragraph{State space.}
The state $S_t \in \mathcal{S}$ should summarize the current ``unlearning
status'' of the model. We build on the context features already used in the
adaptive SISS framework and augment them with the previous action and a notion
of training progress. A simple and effective state representation is
\begin{equation}
    S_t = \big[
    \mathcal{L}_{\text{retain},t},\,
    \mathcal{L}_{\text{forget},t},\,
    \|\nabla_\theta \mathcal{L}_{\text{retain},t}\|,\,
    \|\nabla_\theta \mathcal{L}_{\text{forget},t}\|,\,
    \lambda_{t-1},\,
    t/T
    \big],
\end{equation}
where $L_{\text{retain},t}$ and $L_{\text{forget},t}$ are mini-batch
estimates of the retain and forget losses at step $t$, and
$\|\nabla_\theta \mathcal{L}_{\text{retain},t}\|$,
$\|\nabla_\theta \mathcal{L}_{\text{forget},t}\|$ are the corresponding gradient norms.
This state vector provides a compact, approximately Markovian summary of how
the current model responds to the retain and forget objectives.

\paragraph{Action space.}
The action $A_t \in \mathcal{A}$ is the choice of the forgetting weight
$\lambda_t$ at step $t$:
\begin{equation}
    \mathcal{A}
    = \big\{\lambda \in [\lambda_{\min}, \lambda_{\max}]\big\}
    \subset \mathbb{R},
\end{equation}
where $[\lambda_{\min}, \lambda_{\max}]$ is a predefined admissible range.
In practice, this range can be chosen as a strict subset of $[0,1]$ to avoid
degenerate cases such as $\lambda = 0$ (no forgetting) or $\lambda = 1$
(overly aggressive forgetting).

\paragraph{Transition dynamics.}
Given a state $S_t$ and an action $\lambda_t$, the environment evolves as
follows:
\begin{enumerate}
    \item Sample a mini-batch
    $x_r \sim P_R,\, x_f \sim P_F,\, t' \sim P_T$, and
    $m_{t'} \sim q_{\lambda_t}(m_{t'} \mid x_r, x_f)$ as in the SISS formulation.
    \item Compute the SISS loss $\mathcal{L}_{\text{siss},t}(\lambda_t)$ for this
    mini-batch.
    \item Update the diffusion model parameters by one gradient step:
    \begin{equation}
        \theta_{t+1}
        \leftarrow
        \theta_t - \eta \,\nabla_\theta \mathcal{L}_{\text{siss},t}(\lambda_t),
    \end{equation}
    where $\eta$ is the learning rate.
    \item Recompute the losses and (optionally) their gradient norms under
    $\theta_{t+1}$ to form the next state $S_{t+1}$.
\end{enumerate}
Because of mini-batch sampling and the stochastic mixture
$q_{\lambda_t}(m_{t'} \mid x_r,x_f)$, the transition dynamics
$S_t \mapsto S_{t+1}$ are stochastic.

\paragraph{Reward function.}
Our ultimate objective is to \emph{minimize} the cumulative SISS loss over
the entire unlearning trajectory, while maintaining stability and protecting
the retain distribution. We therefore define a step-wise cost as
\begin{equation}
    C_t(\lambda_t)
    = \mathcal{L}_{\text{siss},t}(\lambda_t)
      + \alpha \big(\lambda_t - \lambda_{t-1}\big)^2,
\end{equation}
where $\alpha > 0$ penalizes abrupt changes in $\lambda$ between consecutive
steps. The corresponding reinforcement learning reward is simply the negative
cost:
\begin{equation}
    R_t
    = r(S_t, \lambda_t)
    = - C_t(\lambda_t)
    = - \mathcal{L}_{\text{siss},t}(\lambda_t)
      - \alpha \big(\lambda_t - \lambda_{t-1}\big)^2.
\end{equation}
Under this definition, maximizing the expected discounted return is equivalent
to minimizing the expected cumulative SISS loss plus a smoothness regularizer
on the sequence of forgetting weights.

\paragraph{Objective.}
Let $\pi_\psi(\lambda \mid S)$ denote a stochastic policy with parameters
$\psi$ that maps states to actions (i.e., selects $\lambda$ based on $S$).
Let $\tau = (S_1, \lambda_1, S_2, \lambda_2, \dots, S_T, \lambda_T)$ denote an
unlearning trajectory induced by this policy and by the stochastic dynamics
of the environment. The reinforcement learning objective is to maximize the
expected discounted return
\begin{equation}
    J(\psi)
    =
    \mathbb{E}_{\tau \sim p_\psi(\tau)}
    \bigg[
        \sum_{t=1}^{T} \gamma^{t-1} R_t
    \bigg]
    =
    \mathbb{E}_{\tau \sim p_\psi(\tau)}
    \bigg[
        -\sum_{t=1}^{T} \gamma^{t-1}
        \Big(
            \mathcal{L}_{\text{siss},t}(\lambda_t)
            + \alpha (\lambda_t - \lambda_{t-1})^2
        \Big)
    \bigg],
\end{equation}
where $\gamma \in (0,1]$ is a discount factor. The optimal dynamic
$\lambda$-selection strategy is then given by a policy
$\pi^\star_\psi$ that maximizes $J(\psi)$.

\subsubsection{Policy parameterization and relation to adaptive \texorpdfstring{$\lambda$}{lambda}}

To keep the reinforcement learning formulation consistent with the adaptive
SISS framework, we parameterize the policy as a Gaussian latent policy
followed by a squashing function that maps the latent variable into the
admissible range of $\lambda$.

Given a state $S_t$, the policy network $f_\psi$ produces the mean and
standard deviation of a Gaussian:
\begin{equation}
    \big(\mu_\psi(S_t),\, \sigma_\psi(S_t)\big)
    = f_\psi(S_t).
\end{equation}
We then sample a latent variable $z_t$ using the reparameterization trick:
\begin{equation}
    z_t = \mu_\psi(S_t) + \xi_t \,\sigma_\psi(S_t),
    \quad
    \xi_t \sim \mathcal{N}(0, 1),
\end{equation}
and obtain the action $\lambda_t$ by squashing $z_t$ into the interval
$[\lambda_{\min}, \lambda_{\max}]$:
\begin{equation}
    \lambda_t
    = \lambda_{\min}
      + (\lambda_{\max} - \lambda_{\min}) \,\sigma(z_t),
\end{equation}
where $\sigma(\cdot)$ is the sigmoid function. This parameterization is
identical in form to the variational encoder for $\lambda$ introduced in
Section~\ref{sec:siss_adaptive}; the key difference lies in the training
objective: instead of minimizing a per-step ELBO, we now optimize the expected
long-term return $J(\psi)$.

To encourage sufficient exploration and to keep the policy close to a
prescribed prior $p(\lambda)$, we additionally include an entropy- or
KL-regularization term. A convenient objective is
\begin{equation}
    \mathcal{L}_{\text{RL}}(\psi)
    = - J(\psi)
      + \beta \,
        \mathbb{E}_{S_t}
        \big[
            D_{\mathrm{KL}}\big(
                \pi_\psi(\lambda \mid S_t) \,\big\|\, p(\lambda)
            \big)
        \big],
\end{equation}
with a regularization weight $\beta > 0$. This yields an
entropy-regularized reinforcement learning objective that is mathematically
analogous to the ELBO used in the variational SISS formulation, but now
extended over entire trajectories instead of single optimization steps.

In practice, we optimize $\mathcal{L}_{\text{RL}}$ using standard
continuous-control policy gradient methods such as Proximal Policy Optimization (PPO).
At each iteration, we roll out several unlearning trajectories under the
current policy $\pi_\psi$, recording states, selected forgetting weights,
rewards, and SISS losses. These trajectories are then used to estimate
advantages and update $\psi$ with a clipped policy-gradient objective. The
diffusion model parameters $\theta$ are updated throughout these rollouts
using standard gradient descent on the SISS loss evaluated at the
policy-selected values of $\lambda_t$. In this way, the policy learns to
choose \emph{dynamic}, state-dependent forgetting weights that optimize the
entire unlearning trajectory, rather than only the instantaneous loss on
each mini-batch.

\subsubsection{Implementation details for PPO-based adaptive \texorpdfstring{$\lambda$}{lambda}}
\label{sec:impl_adaptive_lambda}

We now describe how we instantiate the MDP formulation of
Section~\ref{sec:mdp_unlearning} in practice and how we train the
PPO agent that selects the forgetting weights $\lambda_t$.

\paragraph{Environment and episodes.}
We treat one \emph{environment step} as one gradient update of the diffusion
model parameters $\theta$ on a single mini-batch. An episode corresponds to
$T$ such updates, starting from a fixed pretrained checkpoint $\theta_0$.
Unless otherwise stated, we set $T = 200$ environment steps per episode.
At the beginning of each episode, we reset $\theta \leftarrow \theta_0$
and initialize the previous forgetting weight to a default value
$\lambda_0$ (we use $\lambda_0 = 0.5$ in all experiments).

\paragraph{Mini-batch construction.}
At each environment step $t$, we construct a mini-batch by sampling
a set of ``forget'' examples from $\mathcal{D}_{\text{forget}}$ and a set of
``retain'' examples from $\mathcal{D}_{\text{retain}}$, using the same
sampling strategy as in the original SISS training.
We use a batch size of $B$ samples per step (e.g., $B=64$), with a fixed
ratio between forget and retain samples (e.g., $B/2$ from each set).
Given the current forgetting weight $\lambda_t$ selected by the agent,
we sample $m_t$ from the mixture $q_{\lambda_t}(m_t \mid x,a)$ and compute
the SISS loss $L_{\text{siss},t}(\lambda_t)$ on this mini-batch.

\paragraph{State, action, and reward.}
The state vector $S_t$ at step $t$ is constructed from the current
mini-batch losses and the previous forgetting weight:
\begin{equation}
    S_t = \big[
    \mathcal{L}_{\text{retain},t},\,
    \mathcal{L}_{\text{forget},t},\,
    \|\nabla_\theta \mathcal{L}_{\text{retain},t}\|,\,
    \|\nabla_\theta \mathcal{L}_{\text{forget},t}\|,\,
    \lambda_{t-1},\,
    t/T
    \big],
\end{equation}
where $\mathcal{L}_{\text{retain},t}$ and $\mathcal{L}_{\text{forget},t}$ are the retain and
forget loss components from the SISS objective on the current mini-batch,
and the gradient norms are computed with respect to the diffusion model
parameters $\theta$. In ablations, we also consider a simplified variant
that omits the gradient norms and only uses the scalar losses, which
slightly reduces computation.

The action $A_t$ is the scalar forgetting weight $\lambda_t$ chosen by the
policy. We restrict $\lambda_t$ to an interval
$[\lambda_{\min}, \lambda_{\max}]$ with
$\lambda_{\min} = 0.1$ and $\lambda_{\max} = 0.9$ to avoid degenerate
extremes (no forgetting or excessively aggressive forgetting).

Given the mini-batch SISS loss $\mathcal{L}_{\text{siss},t}(\lambda_t)$, we define
the reward as the negative loss plus a smoothness penalty:
\begin{equation}
    R_t
    = - L_{\text{siss},t}(\lambda_t)
      - \alpha \big(\lambda_t - \lambda_{t-1}\big)^2,
\end{equation}
with $\alpha > 0$ controlling how strongly we penalize large changes in
$\lambda$ between consecutive steps. In our experiments we set
$\alpha = 0.01$. This reward encourages the policy to reduce the SISS loss
while adapting $\lambda_t$ gradually over time.

\paragraph{Policy and value networks.}
Both the policy $\pi_\psi(\lambda \mid S)$ and the value function
$V_\phi(S)$ are implemented as small multi-layer perceptrons (MLPs) that
operate on the state vector $S_t$. Unless otherwise specified, we use two
hidden layers with 64 units each and $\tanh$ activations.

The policy network outputs the mean and log-standard deviation of a
univariate Gaussian:
\begin{equation}
    (\mu_\psi(S_t), \log \sigma_\psi(S_t)) = f_\psi(S_t),
\end{equation}
from which we sample a latent variable $z_t$ using the reparameterization
trick,
\begin{equation}
    z_t = \mu_\psi(S_t) + \xi_t \,\sigma_\psi(S_t),
    \quad
    \xi_t \sim \mathcal{N}(0, 1).
\end{equation}
The forgetting weight is then obtained by squashing $z_t$ to the admissible
range,
\begin{equation}
    \lambda_t
    = \lambda_{\min}
      + (\lambda_{\max} - \lambda_{\min}) \,\sigma(z_t),
\end{equation}
where $\sigma(\cdot)$ is the sigmoid nonlinearity.
The value network $V_\phi(S_t)$ shares the same architecture but outputs
a single scalar estimating the expected return from state $S_t$.

\paragraph{PPO training.}
We optimize the policy and value networks using Proximal Policy Optimization
(PPO). For each PPO update, we collect $N$ episodes of length $T$ by
interacting with the environment as described above, starting each episode
from the same pretrained checkpoint $\theta_0$.
During data collection, the diffusion model parameters $\theta$ are updated
online at every environment step using one gradient step on the SISS loss
$\mathcal{L}_{\text{siss},t}(\lambda_t)$, while the policy parameters $\psi$ and
value parameters $\phi$ are kept fixed. The resulting trajectories
$\{(S_t, \lambda_t, R_t, S_{t+1})\}$ are stored for PPO.

Given the collected trajectories, we compute discounted returns and
generalized advantage estimates (GAE) with discount factor
$\gamma = 0.99$ and GAE parameter $\lambda_{\mathrm{GAE}} = 0.95$.
We then perform several epochs of PPO updates on mini-batches of the stored
transitions, using the clipped objective with ratio
$r_t(\psi) = \pi_\psi(\lambda_t \mid S_t) /
             \pi_{\psi_{\text{old}}}(\lambda_t \mid S_t)$
and clipping parameter $\epsilon = 0.2$. We use Adam with a learning rate
of $3 \times 10^{-4}$ for both actor and critic, and include a small
entropy bonus to encourage exploration.

After PPO training has converged, we fix the learned policy
$\pi_\psi(\lambda \mid S)$ and perform a final unlearning run, during which
$\theta$ is updated using SISS loss with $\lambda_t$ chosen by the trained
policy at each step. All unlearning metrics reported in the main text are
computed from this final run.

\section{Experiments}
\subsection*{Experiment A}
\begin{figure}[h!]
    \centering
    \includegraphics[scale=0.2]{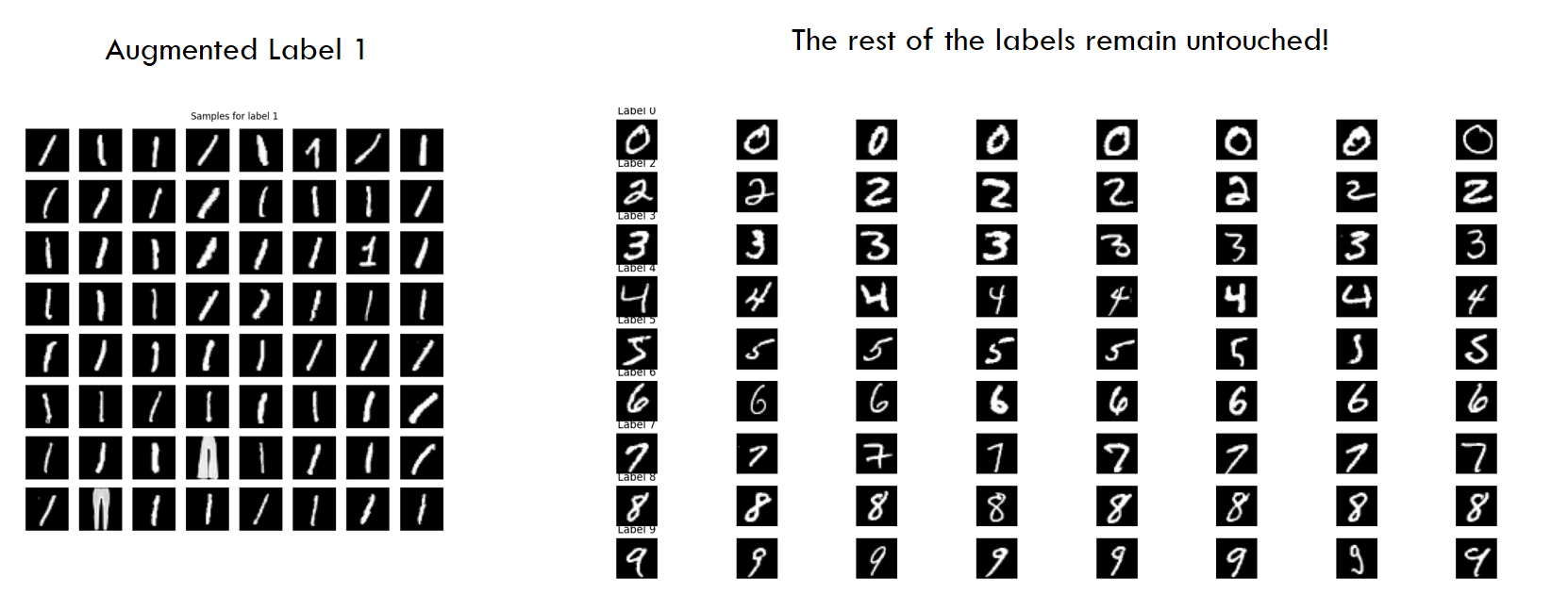} 
    \caption{The Augmented-Mnist Dataset is depicted in this image. }
    \label{fig:augmented_dataset}
\end{figure}

To illustrate practical behavior, we construct an augmented MNIST dataset in which class “\textbf{1}” is contaminated with \textit{Fashion-MNIST} trousers in a $1:10$ ratio. A conditional diffusion model trained on these data learns hybrid semantics for the label “\textbf{1}” (refer to Figure~\ref{fig:augmented_dataset}). We train: a baseline model, an SISS unlearning, our adaptive-$\lambda$ method. It is also worth mentioning that we use EMA smoothing and Flipping to make the Baseline DDPM model be able to generate trousers as well conditional on label $1$.

Across repeated runs, qualitative samples and quantitative metrics: KID, FID, SSIM, and IS; show that our framework more effectively removes trouser content while restoring clean digit structure, outperforming SISS consistently (please refer to Table~\ref{tab:metrics-comparison} \& Appendix). Additional CIFAR-10 and ImageNet results will be included in the final paper.


\begin{table}[htbp]
\centering
\caption{Comparison of Metrics Across Methods (Mean $\pm$ Std)}
\label{tab:metrics-comparison}
\begin{tabular}{l|ccc|ccc}
\toprule
\multirow{2}{*}{\textbf{Metric}} &
  \multicolumn{3}{c|}{\textbf{Forget Set}} &
  \multicolumn{3}{c}{\textbf{Retain Set}} \\
\cmidrule(lr){2-4} \cmidrule(lr){5-7}
& Baseline & Dynamic & SISS & Baseline & Dynamic & SISS \\
\midrule
FID  & $145.03 \pm 12.5$ & $214.83 \pm 18.0$ & $243.51 \pm 20.5$ & $32.92 \pm 2.8$ & $39.41 \pm 3.2$ & $62.67 \pm 5.1$ \\
KID  & $0.93 \pm 0.21$   & $1.72 \pm 0.19$   & $0.83 \pm 0.12$   & $0.06 \pm 0.01$ & $0.06 \pm 0.01$ & $0.16 \pm 0.02$ \\
IS   & $2.20 \pm 0.16$   & $1.69 \pm 0.06$   & $1.89 \pm 0.07$   & $1.75 \pm 0.12$ & $1.73 \pm 0.11$ & $1.72 \pm 0.10$ \\
SSIM & $0.65 \pm 0.08$   & $0.70 \pm 0.17$   & $0.86 \pm 0.11$   & $0.38 \pm 0.03$ & $0.42 \pm 0.03$ & $0.29 \pm 0.02$ \\
\bottomrule
\end{tabular}
\end{table}


For the MNIST dataset, in the tasks that we defined, for the FID score metrics (the lower the score, the better), in the retain set our model has a lower score vs. the SISS framework with respect to the baseline. This is also visible in per-class evaluations. In KID metrics (the lower the better as well), our model is getting much lower KID vs. SISS with respect to the baseline model, and it is having much higher KID with respect to the baseline and SISS as we want our model not to perform well on the forget set! We can also see how much we are close to baseline metrics per class in general. The same things apply to SSIM metrics; however, there is not much difference in IS scores.

\begin{figure}[ht]
    \centering
    \includegraphics[scale=0.4]{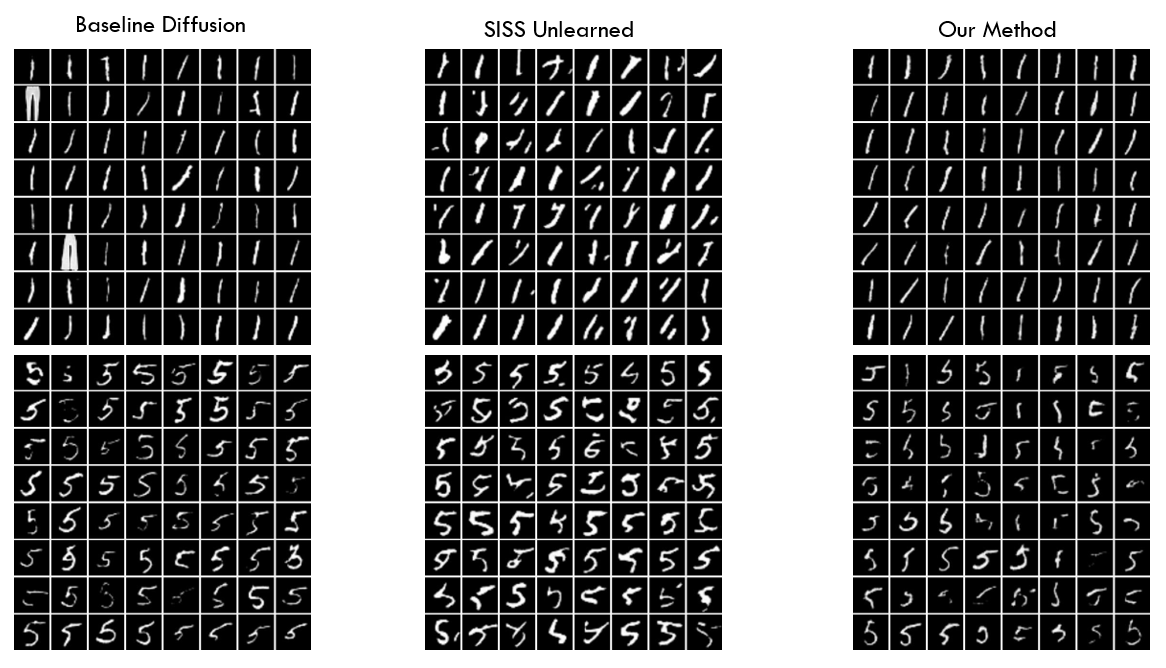}
    \caption{Visualization of samples generated by each model: baseline (left), SISS 
    (middle), and our method (right).}
    \label{fig:experiment_a_model_perfs}
\end{figure}

Furthermore, the qualitative comparison in Figure~\ref{fig:experiment_a_model_perfs}
illustrates the visual differences between the generated samples. The baseline model is 
shown on the left, SISS in the middle, and our proposed method on the right. In our 
experiments, the samples produced by our method exhibit clearer separation between 
retained and forgotten concepts compared to the SISS framework.

\section{Conclusion}
\label{sec:conclusion}

In this work, we addressed the critical challenge of balancing the competing objectives of model fidelity and unlearning efficacy within the Stochastic Sampling Importance Sampling (SISS) framework for machine unlearning in Diffusion Models\cite{alberti2025dataunlearning}. We identified the primary limitation of existing methods as the use of a static $\lambda$ hyperparameter, which fails to account for the dynamic, sample-dependent nature of the unlearning process. To overcome this, we introduced \textbf{the Adaptive-$\lambda$ SISS} framework, which leverages Variational Inference to learn an optimal, context-dependent distribution over $\lambda$. By conditioning $\lambda$ on a feature vector derived from the current \textit{retain and forget losses}, our approach enables the model to dynamically adjust the penalty applied to the forget objective during training.

Our empirical analysis demonstrated that the Adaptive-$\lambda$ approach achieves superior performance across challenging concept erasure tasks. Specifically, the dynamic approach significantly improved the unlearning metrics (e.g., FID, KID) for the Forget Set, confirming greater erasure efficacy, while maintaining or even improving the generative fidelity metrics (e.g., IS, SSIM) on the Retain Set compared to the Static-$\lambda$ baseline. This substantiates the core hypothesis: a learned, adaptive loss balancing mechanism yields a more stable and effective unlearned model, successfully mitigating the common trade-off between forgetting and retaining capabilities.

We also extended this dynamic loss balancing to a more sophisticated, sequential decision-making process by formalizing the unlearning task as a Reinforcement Learning (RL) problem. This framework will model the loss balancing as a Markov Decision Process (MDP), allowing an agent to learn a policy that maximizes a reward function tied to both retention and forgetting metrics over an entire unlearning epoch, promising an even greater degree of control and performance stability.


\appendix
\newpage

\section*{Appendix A}

\begin{figure}[h!]
\centering

\fbox{%
\begin{minipage}{0.95\columnwidth}
    \vspace{0.6em}

    \centering
    \textbf{Algorithm 2: Adaptive-$\lambda$ SISS Unlearning for DDPMs}
    \vspace{0.5em}
    \hrule
    \vspace{0.5em}

    \footnotesize  

    \begin{algorithmic}[1]

    \Require Dataset $\mathcal{D}$, Forget set $\mathcal{F}$, Retain set $\mathcal{R}$,
    Model $\varepsilon_\theta$, Inference Net $\varepsilon_\phi$, noise $q$,
    learning rate $\eta$, scaling factor $s$, KL-weight $\beta$, epochs $E$

    \State Initialize $\theta$, $\phi$, prior $p(\lambda)$

    \For{$e = 1$ to $E$}

        \State Sample $x \sim \mathcal{R}$, $a \sim \mathcal{F}$, $t \sim P_T$

        \State \textbf{Base Losses:}
        \Statex \hspace{\algorithmicindent}
        $\mathcal{L}_{\mathrm{ret}}^{\mathrm{base}}
            = \left\|\frac{m_t - \gamma_t x}{\sigma_t}
            - \varepsilon_\theta(m_t,t)\right\|^2$
        \Statex \hspace{\algorithmicindent}
        $\mathcal{L}_{\mathrm{forget}}^{\mathrm{base}}
            = \left\|\frac{m_t - \gamma_t a}{\sigma_t}
            - \varepsilon_\theta(m_t,t)\right\|^2$

        \State \textbf{Context:}
        $v = \Big[\mathcal{L}_{\mathrm{ret}}^{\mathrm{base}},
             \mathcal{L}_{\mathrm{forget}}^{\mathrm{base}},
             ||\nabla_\theta\mathcal{L}_{\mathrm{ret}}^{\mathrm{base}}||,
             ||\nabla_\theta\mathcal{L}_{\mathrm{forget}}^{\mathrm{base}}||\Big]$

        \State \textbf{Adaptive $\lambda$ (Reparam):}
        \Statex \hspace{\algorithmicindent}
        $(\mu_\phi, \sigma_\phi) = \varepsilon_\phi(v)$;
        sample $\xi \sim \mathcal{N}(0,I)$,
        $z = \mu_\phi + \xi\sigma_\phi$,
        $\lambda = \sigma(z)$

        \State Sample latent:
        $m_t \sim q_\lambda(m_t)
        = (1-\lambda)q(m_t|x) + \lambda q(m_t|a)$

        \State \textbf{Weighted Losses:}
        \Statex \hspace{\algorithmicindent}
        $\mathcal{L}_{\mathrm{ret}}(\lambda)
        = \frac{|\mathcal{D}|}{|\mathcal{R}|}
        \frac{q(m_t|x)}{q_\lambda(m_t)}
        \mathcal{L}_{\mathrm{ret}}^{\mathrm{base}}$

        \Statex \hspace{\algorithmicindent}
        $\mathcal{L}_{\mathrm{forget}}(\lambda)
        = \frac{|\mathcal{D}|}{|\mathcal{F}|}
        \frac{q(m_t|a)}{q_\lambda(m_t)}
        \mathcal{L}_{\mathrm{forget}}^{\mathrm{base}}$

        \State \textbf{Total:}
        $\mathcal{L}_{\mathrm{SISS}}
        = \mathcal{L}_{\mathrm{ret}}(\lambda)
        - (1+s)\mathcal{L}_{\mathrm{forget}}(\lambda)$

        \State \textbf{ELBO:}
        $\mathcal{L}_{\mathrm{ELBO}}
        = \mathcal{L}_{\mathrm{SISS}}
        + \beta D_{\mathrm{KL}}(\mathcal{N}(\mu_\phi,\sigma_\phi) \,\|\, p(\lambda))$

        \State Update:
        $\theta \leftarrow \theta - \eta\nabla_\theta\mathcal{L}_{\mathrm{ELBO}}$,
        $\phi \leftarrow \phi - \eta\nabla_\phi\mathcal{L}_{\mathrm{ELBO}}$

    \EndFor

    \end{algorithmic}

    \vspace{0.5em}
\end{minipage}
} 

\caption{Adaptive-$\lambda$ SISS Unlearning using a VAE-like inference network for dynamic $\lambda$.}
\label{alg:adaptive_siss_corrected}

\end{figure}

\newpage
\section*{Appendix B}

\newcommand{\imgscale}{0.22}
\newcommand{\gap}{0.3cm}

\begin{figure}[ht]
    \centering

    \begin{minipage}{0.48\linewidth}
        \centering
        \textbf{Average Metrics}
        \vspace{0.3cm}

        \includegraphics[scale=\imgscale]{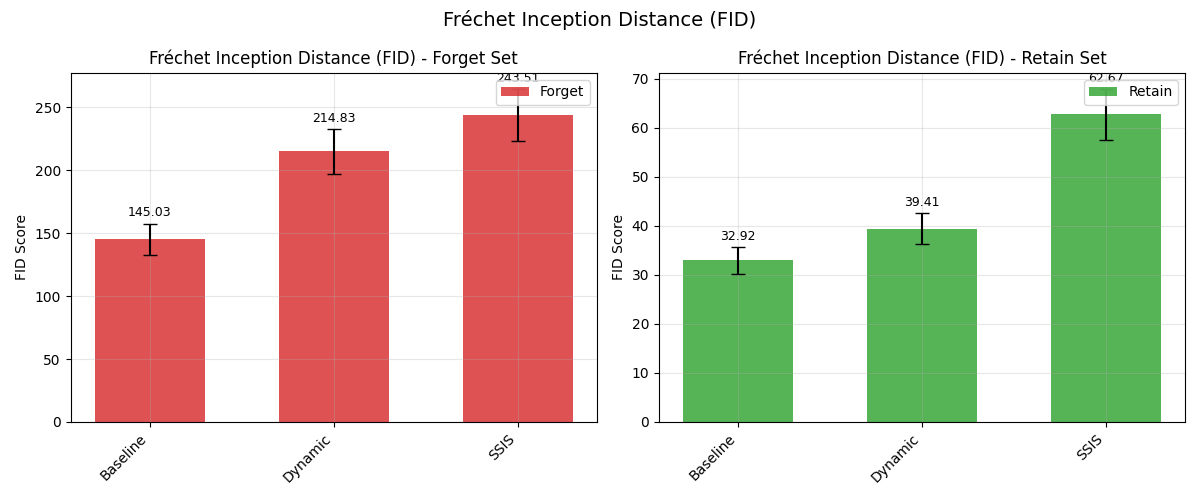}
        \hspace{\gap}
        \includegraphics[scale=\imgscale]{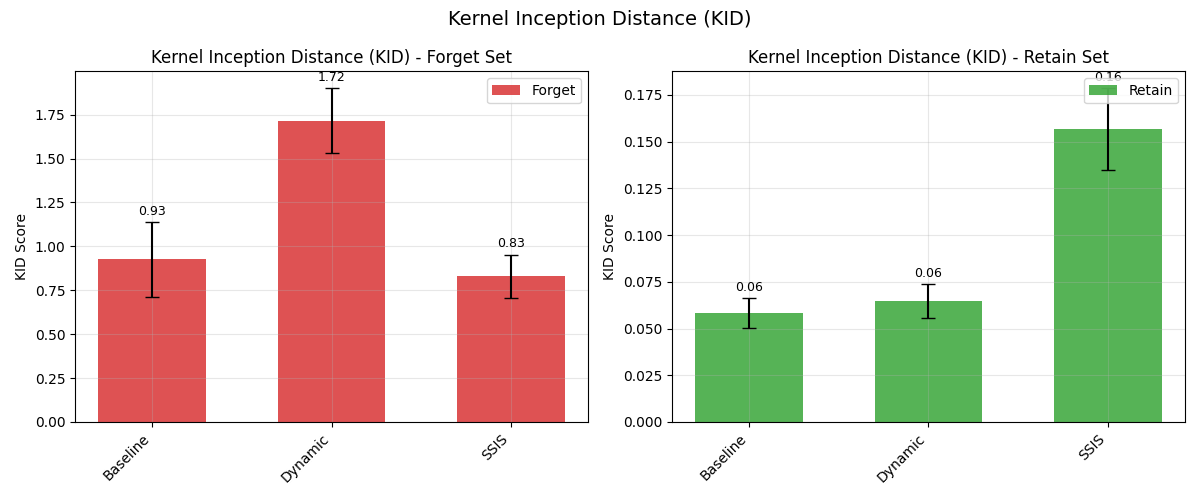}

        \vspace{0.4cm}

        \includegraphics[scale=\imgscale]{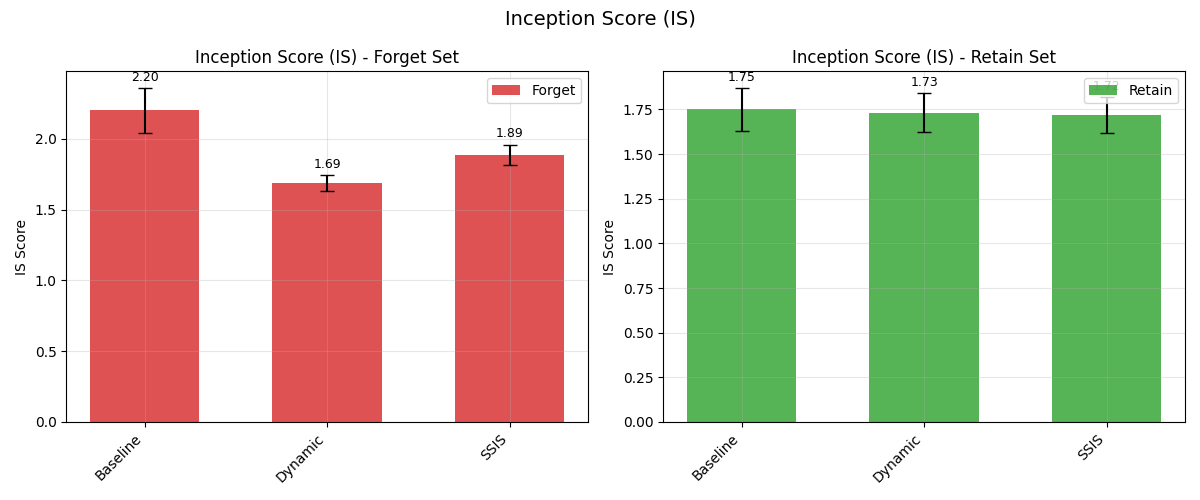}
        \hspace{\gap}
        \includegraphics[scale=\imgscale]{figs/is.png}

        \label{fig:aaaaaaaaaaa}
    \end{minipage}
    \hfill
    \begin{minipage}{0.48\linewidth}
        \centering
        \textbf{Per-Class Metrics}
        \vspace{0.3cm}

        \includegraphics[scale=\imgscale]{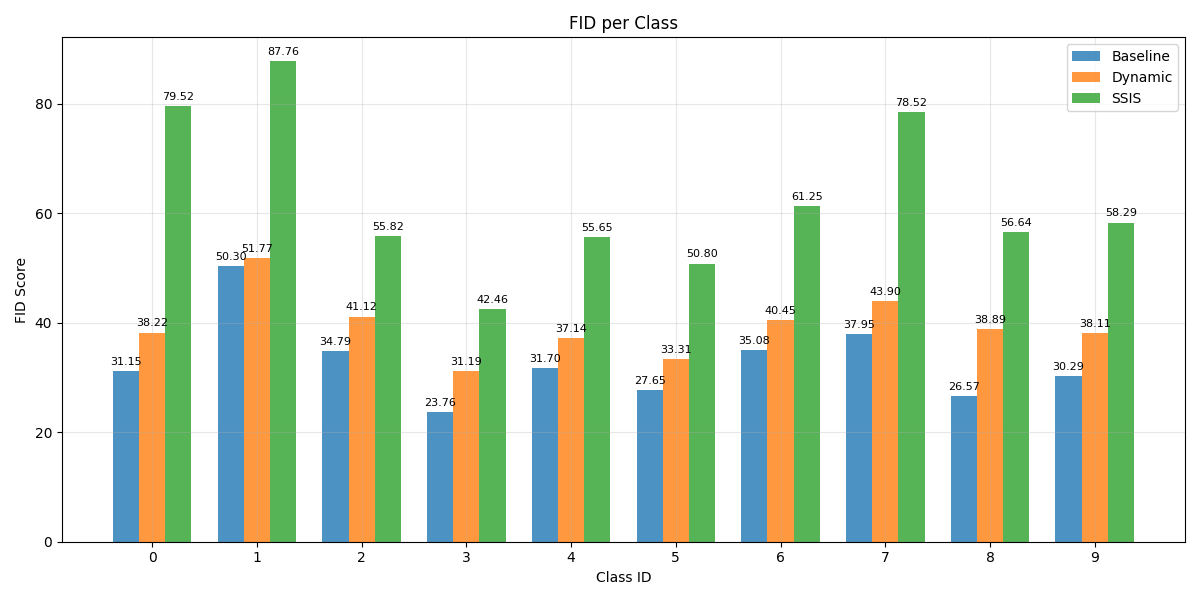}
        \hspace{\gap}
        \includegraphics[scale=\imgscale]{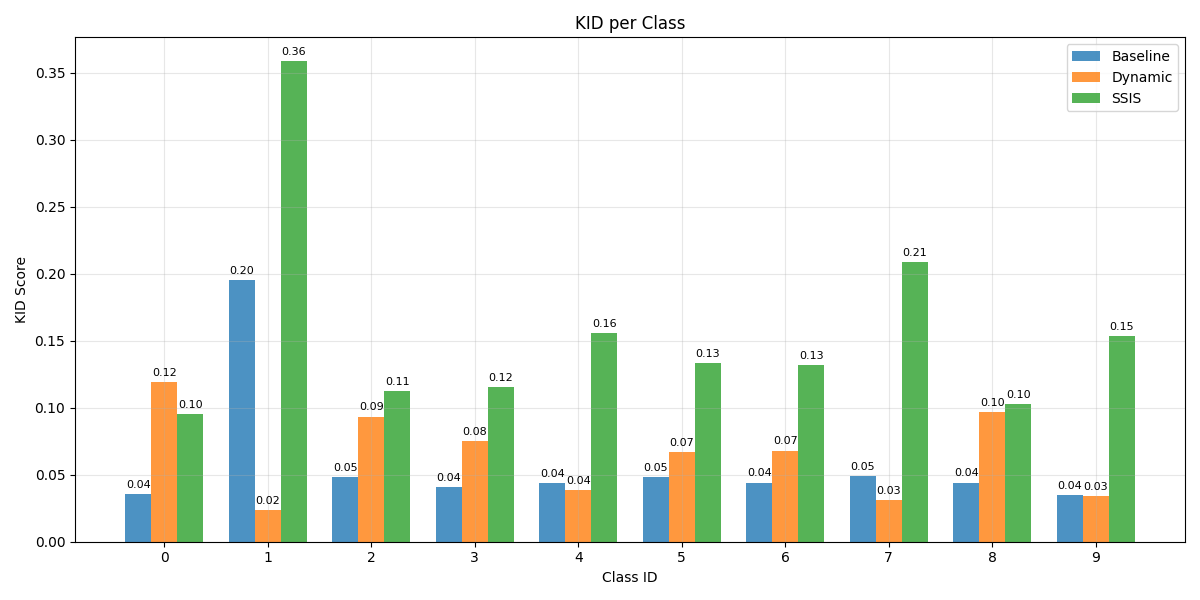}

        \vspace{0.4cm}

        \includegraphics[scale=\imgscale]{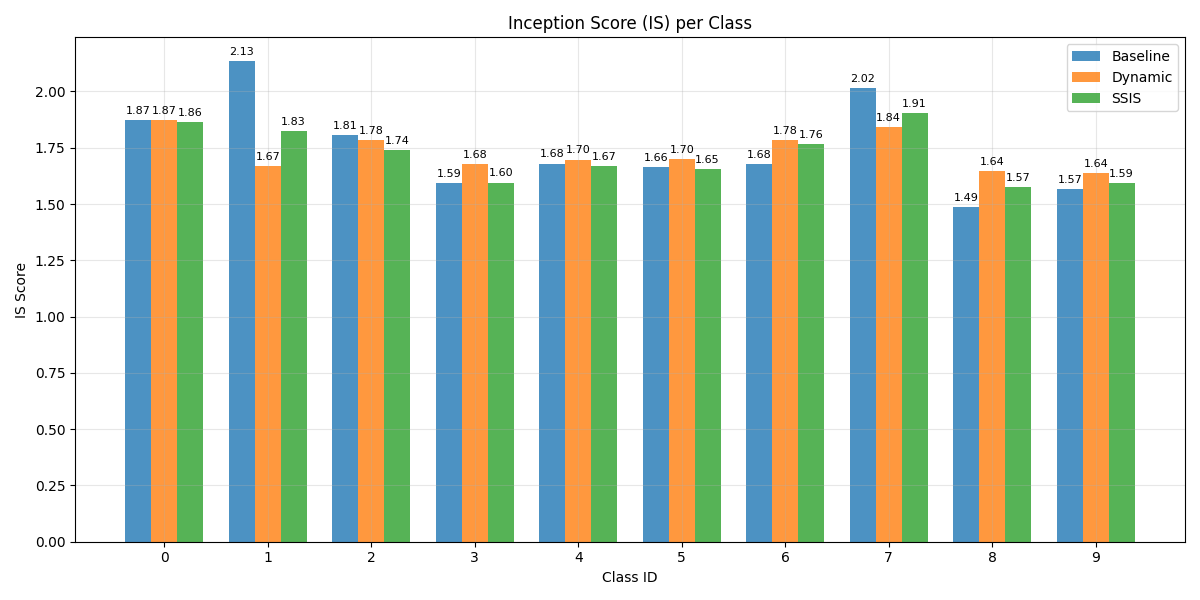}
        \hspace{\gap}
        \includegraphics[scale=\imgscale]{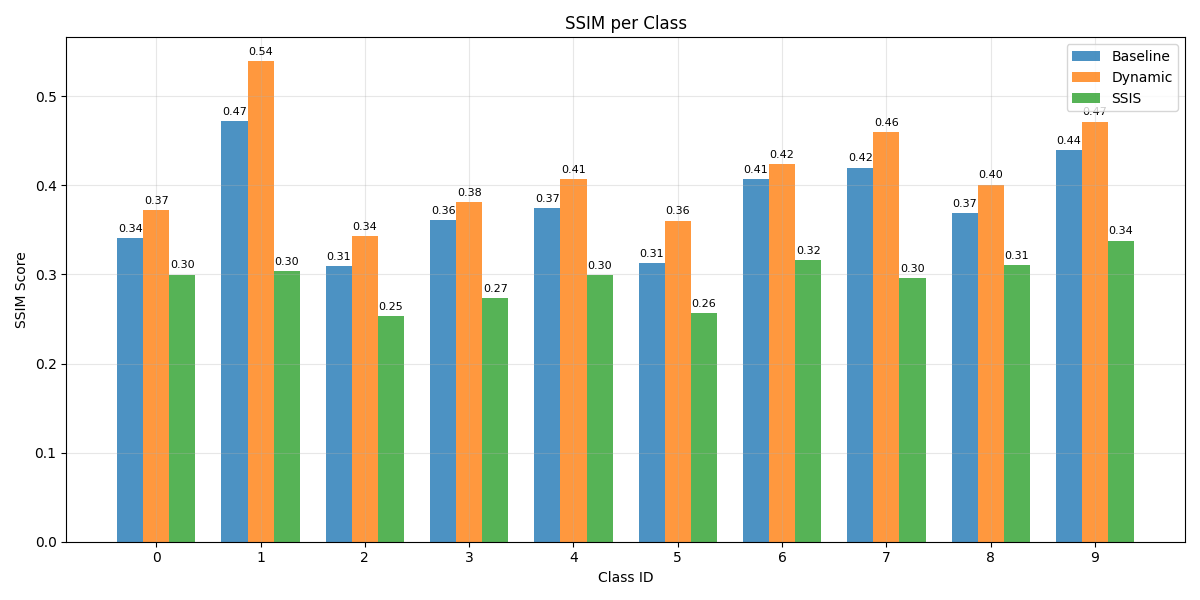}

        \label{fig:bbbbbbbbbbbbb}
    \end{minipage}

    \caption{MNIST experiment Average and Per-Class metrics displayed side-by-side in a single row.}
\end{figure}

\section*{Appendix C: RL Experiment details}
\label{sec:impl_adaptive_lambda_options}

We now describe two practical strategies for training the reinforcement
learning agent that selects the forgetting weights $\lambda_t$ in the MDP
formulation of Section~\ref{sec:mdp_unlearning}. Both are based on
continuous-control algorithms with stochastic policies; in the main
experiments we use PPO, and we additionally verify that Soft Actor--Critic
(SAC) is also applicable to this setting.

Throughout this section, we treat one \emph{environment step} as one gradient
update of the diffusion model parameters $\theta$ on a single mini-batch, and
we use the same state, action, and reward definitions as in
Section~\ref{sec:mdp_unlearning}. In particular, the state is
\begin{equation}
    S_t = \big[
    L_{\text{retain},t},\,
    L_{\text{forget},t},\,
    \|\nabla_\theta L_{\text{retain},t}\|,\,
    \|\nabla_\theta L_{\text{forget},t}\|,\,
    \lambda_{t-1},\,
    t/T
    \big],
\end{equation}
the action is the scalar forgetting weight
$\lambda_t \in [\lambda_{\min}, \lambda_{\max}]$ with
$\lambda_{\min} = 0.1$ and $\lambda_{\max} = 0.9$, and the reward is
\begin{equation}
    R_t
    = - L_{\text{siss},t}(\lambda_t)
      - \alpha \big(\lambda_t - \lambda_{t-1}\big)^2,
\end{equation}
with $\alpha = 0.01$ unless otherwise noted.

\paragraph{Mini-batch construction.}
At each environment step $t$, we construct a mini-batch of size $B$ by
sampling from the retain set $\mathcal{D}_{\text{retain}}$ and forget set
$\mathcal{D}_{\text{forget}}$. Unless otherwise specified, we use $B=64$ with
a balanced split between retain and forget samples. Given the current
forgetting weight $\lambda_t$ selected by the policy, we sample
$m_t \sim q_{\lambda_t}(m_t \mid x,a)$ as in the SISS formulation and compute
the mini-batch SISS loss $L_{\text{siss},t}(\lambda_t)$, which is used both
to update $\theta$ and to compute the reward $R_t$.

\paragraph{Policy and value architectures.}
The policy $\pi_\psi(\lambda \mid S)$ and value function $V_\phi(S)$ are
implemented as small multi-layer perceptrons (MLPs) operating on the state
vector $S_t$. In all experiments, we use two hidden layers with 64 units each
and $\tanh$ activations. The policy network outputs the mean and log-standard
deviation of a univariate Gaussian,
\begin{equation}
    (\mu_\psi(S_t), \log \sigma_\psi(S_t)) = f_\psi(S_t),
\end{equation}
from which we sample a latent variable $z_t$ using the reparameterization
trick,
\begin{equation}
    z_t = \mu_\psi(S_t) + \xi_t \,\sigma_\psi(S_t),
    \quad
    \xi_t \sim \mathcal{N}(0, 1),
\end{equation}
and squash it into the admissible range via
\begin{equation}
    \lambda_t
    = \lambda_{\min}
      + (\lambda_{\max} - \lambda_{\min}) \,\sigma(z_t),
\end{equation}
where $\sigma(\cdot)$ is the sigmoid function.
The value network $V_\phi(S_t)$ shares the same architecture but outputs a
single scalar.

\subparagraph{Option 1: Episodic unlearning with parameter resets.}
\label{sec:impl_option1}

In the first implementation strategy, we construct a ``clean'' MDP by
resetting the diffusion model to a fixed pretrained checkpoint at the start
of every episode. One episode consists of $T$ environment steps, each
corresponding to one mini-batch update of $\theta$.

\begin{itemize}
    \item \textbf{Episode initialization.}
    At the beginning of each episode, we reset
    $\theta \leftarrow \theta_0$, where $\theta_0$ is the pretrained model
    before unlearning, and initialize the previous forgetting weight to a
    default value $\lambda_0$ (we use $\lambda_0 = 0.5$). We also perform a
    single warm-up forward pass to compute the initial losses and build
    $S_1$.

    \item \textbf{Within-episode interaction.}
    For each step $t = 1,\dots,T$ within an episode:
    \begin{enumerate}
        \item Given the current state $S_t$, the policy samples
        $\lambda_t \sim \pi_\psi(\cdot \mid S_t)$.
        \item We construct a mini-batch from
        $\mathcal{D}_{\text{retain}}$ and $\mathcal{D}_{\text{forget}}$,
        sample $m_t \sim q_{\lambda_t}(m_t \mid x,a)$, and compute the
        SISS loss $L_{\text{siss},t}(\lambda_t)$ on this mini-batch.
        \item We update the diffusion model parameters by one gradient step:
        \begin{equation}
            \theta_{t+1}
            \leftarrow
            \theta_t - \eta \,\nabla_\theta L_{\text{siss},t}(\lambda_t),
        \end{equation}
        using Adam with learning rate $\eta$ (e.g., $\eta = 10^{-4}$).
        \item We compute the reward $R_t$ from the SISS loss and smoothness
        penalty, and rebuild the next state $S_{t+1}$ by recomputing
        $L_{\text{retain},t+1}$ and $L_{\text{forget},t+1}$ (and their
        gradient norms) under $\theta_{t+1}$.
        \item We store the transition
        $(S_t, \lambda_t, R_t, S_{t+1})$ as a sample for PPO.
    \end{enumerate}

    \item \textbf{PPO updates.}
    After collecting $N$ such episodes (each of length $T$), we perform
    PPO updates on the policy and value networks. We compute discounted
    returns and generalized advantage estimates (GAE) with discount factor
    $\gamma = 0.99$ and GAE parameter $\lambda_{\mathrm{GAE}} = 0.95$, and
    optimize the clipped PPO objective with clipping parameter
    $\epsilon = 0.2$. We use Adam with learning rate $3 \times 10^{-4}$ for
    both actor and critic and add a small entropy bonus to encourage
    exploration. The unlearning model parameters $\theta$ used during data
    collection are discarded at the end of each episode, and do not directly
    carry over across episodes.
\end{itemize}

In this option, the unlearning process is simulated repeatedly from the same
starting point $\theta_0$ in order to train the policy. After convergence of
PPO, we fix $\pi_\psi$ and perform a final unlearning run starting from
$\theta_0$, during which $\theta$ is updated using SISS with
$\lambda_t$ chosen by the trained policy at each step. All reported
unlearning metrics for Option~1 are computed from this final run.

\subparagraph{Option 2: Online unlearning in a single long run.}
\label{sec:impl_option2}

The second strategy uses a single, long unlearning trajectory, without
resetting the diffusion model parameters between episodes. This option is
closer to the practical deployment setting, and we use it as the primary
setup in our main experiments.

\begin{itemize}
    \item \textbf{Single long trajectory.}
    We start from the pretrained model $\theta_0$ and run unlearning for
    $T_{\text{total}}$ environment steps (e.g.,
    $T_{\text{total}} \in [10^3, 10^4]$ depending on the dataset). At each
    step $t$, we:
    \begin{enumerate}
        \item Build the state $S_t$ from the current losses and previous
        forgetting weight.
        \item Sample an action $\lambda_t \sim \pi_\psi(\cdot \mid S_t)$.
        \item Sample a mini-batch and compute
        $L_{\text{siss},t}(\lambda_t)$.
        \item Update $\theta$ by one gradient step on
        $L_{\text{siss},t}(\lambda_t)$.
        \item Compute the reward $R_t$ and the next state $S_{t+1}$.
        \item Store the transition $(S_t, \lambda_t, R_t, S_{t+1})$ in a
        buffer of recent interactions.
    \end{enumerate}

    \item \textbf{Periodic PPO updates.}
    Every $M$ environment steps (e.g., $M = 200$), we pause data collection
    and perform PPO updates using the most recent window of transitions
    (e.g., the last $N_{\text{steps}} = N \cdot T$ steps, matching the
    effective batch size used in Option~1). We reuse the same hyperparameters
    as in Option~1: $\gamma = 0.99$,
    $\lambda_{\mathrm{GAE}} = 0.95$, $\epsilon = 0.2$, and Adam with learning
    rate $3 \times 10^{-4}$.
\end{itemize}

In this option, the diffusion model parameters $\theta$ continuously evolve
over time, and the same trajectory is used both to train the policy and to
perform actual unlearning. Despite the non-stationarity induced by the
changing model, PPO remains stable in practice and converges to a policy that
selects meaningful dynamic forgetting weights. In all main experiments, we
start from this online Option~2 and additionally report results for Option~1
as an ablation to isolate the effect of the episodic resets.

\subparagraph{Using Soft Actor--Critic (SAC).}
\label{sec:impl_sac}

Because the action space for $\lambda_t$ is continuous and low-dimensional,
off-policy actor--critic methods designed for continuous control are also
applicable. In particular, we found that Soft Actor--Critic (SAC) can be used
as a drop-in replacement for PPO in both Option~1 and Option~2, with minor
modifications:

\begin{itemize}
    \item The policy $\pi_\psi(\lambda \mid S)$ retains the same squashed
    Gaussian parameterization as above, but is now trained to \emph{maximize}
    both expected return and an entropy term, following the SAC objective.
    \item We introduce two Q-functions $Q_{\theta_1}(S,\lambda)$ and
    $Q_{\theta_2}(S,\lambda)$ implemented as MLPs with the same architecture
    as the value network, and train them with Bellman backups on transitions
    $(S_t, \lambda_t, R_t, S_{t+1})$ stored in a replay buffer.
    \item The actor is updated by minimizing the SAC policy loss
    \begin{equation}
        \mathcal{L}_{\text{actor}}(\psi)
        =
        \mathbb{E}_{S_t \sim \mathcal{D},\, \lambda_t \sim \pi_\psi}
        \big[
            \alpha_{\text{ent}} \log \pi_\psi(\lambda_t \mid S_t)
            - \min_{i=1,2} Q_{\theta_i}(S_t, \lambda_t)
        \big],
    \end{equation}
    where $\alpha_{\text{ent}}$ is the entropy temperature and
    $\mathcal{D}$ denotes the replay buffer.
    \item In Option~1, the replay buffer is filled by repeatedly simulating
    episodes from the reset model $\theta_0$; in Option~2, the buffer is
    populated online during the single long unlearning run.
\end{itemize}

Empirically, SAC achieves similar qualitative behavior to PPO in terms of
learning dynamic, state-dependent forgetting weights, and sometimes converges
in fewer environment steps due to its off-policy nature and replay buffer.
However, PPO is simpler to integrate into our existing training loop and is
used as the default algorithm in the main results, while SAC is included as
an alternative in ablation studies.


\begin{thebibliography}{9}

\bibitem{kingma2013auto}
Kingma, D. P., \& Welling, M. (2013). Auto-Encoding Variational Bayes. \emph{arXiv preprint arXiv:1312.6114}.

\bibitem{rezende2014stochastic}
Rezende, D. J., Mohamed, S., \& Wierstra, D. (2014). Stochastic Backpropagation and Approximate Inference in Deep Generative Models. \emph{arXiv preprint arXiv:1401.4082}.

\bibitem{sohl2015deep}
Sohl-Dickstein, J., Weiss, E., Mané, D., Allec, R., \& Ermon, S. (2015). Deep Unsupervised Learning using Nonequilibrium Thermodynamics. \emph{arXiv preprint arXiv:1503.03585}.

\bibitem{ho2020denoising}
Ho, J., Jain, A., \& Abbeel, P. (2020). Denoising Diffusion Probabilistic Models. \emph{arXiv preprint arXiv:2006.11239}.

\bibitem{song2019generative}
Song, Y., \& Ermon, S. (2019). Generative Modeling by Estimating Gradients of the Data Distribution. \emph{arXiv preprint arXiv:1907.05600}.

\bibitem{alberti2025dataunlearning}
Alberti, S., Hasanaliyev, K., Shah, M., \& Ermon, S. (2025).
Data Unlearning in Diffusion Models.
\emph{arXiv preprint arXiv:2503.01034}.

\bibitem{chen2024score}
Chen, T., Zhang, S., \& Zhou, M. (2024). Score Forgetting Distillation: A Swift, Data-Free Method for Machine Unlearning in Diffusion Models. \emph{arXiv preprint arXiv:2409.11219}.

\bibitem{song2020scorebased}
Song, Y., Chen, J., \& Ermon, S. (2020). Score-Based Generative Modeling through Stochastic Differential Equations. \emph{arXiv preprint arXiv:2011.13456}.

\bibitem{li2024machine_unlearn}
Li, G., Hsu, H., Chen, C.-F. (Richard), \& Marculescu, R. (2024).
Machine Unlearning for Image-to-Image Generative Models.
\emph{arXiv preprint arXiv:2402.00351}.
\bibitem{cao2015towards}
Cao, Y., \& Yang, J. (2015). Towards making systems forget with machine unlearning. In \emph{2015 IEEE Symposium on Security and Privacy} (pp. 463-480). IEEE.

\bibitem{ginart2019making}
Ginart, A., Guan, M., Valiant, G., \& Zou, J. (2019). Making AI forget you: Data deletion in machine learning. In \emph{Advances in Neural Information Processing Systems} (NeurIPS).

\bibitem{guo2020certified}
Guo, C., Goldstein, T., Hannun, A., \& van der Maaten, L. (2020). Certified data removal from machine learning models. In \emph{International Conference on Machine Learning} (ICML) (pp. 3832-3842). PMLR.

\bibitem{golatkar2020eternal}
Golatkar, A., Achille, A., \& Soatto, S. (2020). Eternal sunshine of the spotless net: Selective forgetting in deep networks. In \emph{Proceedings of the IEEE/CVF Conference on Computer Vision and Pattern Recognition} (CVPR) (pp. 9304-9312).

\bibitem{bourtoule2021machine}
Bourtoule, L., Chandrasekaran, V., Choquette-Choo, C. A., et al. (2021). Machine unlearning. In \emph{2021 IEEE Symposium on Security and Privacy (SP)} (pp. 141-159). IEEE.

\bibitem{sekhari2021remember}
Sekhari, A., Acharya, J., Kamath, G., \& Suresh, A. T. (2021). Remember what you want to forget: Algorithms for machine unlearning. In \emph{Advances in Neural Information Processing Systems} (NeurIPS) (Vol. 34, pp. 18075-18086).

\bibitem{baumhauer2022machine}
Baumhauer, T., Schramowski, P., \& Kersting, K. (2022). Machine unlearning of features and labels. \emph{arXiv preprint arXiv:2108.11577}.

\bibitem{gandikota2023erasing}
Gandikota, S., Materzynska, J., Fowlkes, C. C., \& Shafahi, A. (2023). Erasing concepts from diffusion models. In \emph{Proceedings of the IEEE/CVF International Conference on Computer Vision} (ICCV) (pp. 2426-2436).

\bibitem{fan2023salun}
Fan, C., Liu, J., Atlas, M., Huang, Z., Li, C., \& Geng, X. (2023). Salun: Empowering machine unlearning via gradient-based weight saliency in both image classification and generation. \emph{arXiv preprint arXiv:2310.12508}.



\bibitem{gupta2021adaptive}
Gupta, V., Tian, C., Zhang, Y., Oprea, A., Rengarajan, A., \& Reiter, K. (2021). Adaptive machine unlearning. In \emph{Advances in Neural Information Processing Systems} (NeurIPS) (Vol. 34, pp. 16319-16330).

\bibitem{song2021scorebased}
Song, Y., Sohl-Dickstein, J., Kingma, D. P., Kumar, A., Ermon, S., \& Poole, B. (2021). Score-based generative modeling through stochastic differential equations. In \emph{International Conference on Learning Representations} (ICLR).

\bibitem{thudi2022unrolling}
Thudi, A., De, S., Peebles, B., \& Pananjady, A. (2022). Unrolling SGD: Understanding factors influencing machine unlearning. In \emph{2022 IEEE European Symposium on Security and Privacy (EuroS\&P)} (pp. 303-319). IEEE.

\bibitem{tarun2023fast}
Tarun, A. K., Chundawat, V. S., Mandal, M., \& Kankanhalli, M. (2023). Fast yet effective machine unlearning. \emph{IEEE Transactions on Neural Networks and Learning Systems}.

\bibitem{kumari2023ablating}
Kumari, N., Zhang, B., Zhang, S., Shechtman, E., Zhang, R., Zhu, J.-Y., \& Agrawala, M. (2023). Ablating concepts in text-to-image diffusion models. In \emph{Proceedings of the IEEE/CVF International Conference on Computer Vision} (ICCV) (pp. 22691-22702).

\bibitem{orgad2024editing}
Orgad, H., Kawar, B., \& Belinkov, Y. (2024). Editing implicit assumptions in text-to-image diffusion models. In \emph{Proceedings of the IEEE/CVF Conference on Computer Vision and Pattern Recognition} (CVPR) (pp. 8655-8665).

\bibitem{kurmanji2024towards}
Kurmanji, M., Triantafillou, E., Hayes, J., \& Triantafillou, E. (2024). Towards unbounded machine unlearning. \emph{arXiv preprint arXiv:2302.09880}.

\bibitem{zhang2024survey}
Zhang, S., Fan, L., Liu, H., Li, J., Dong, Y., \& Xu, K. (2024). Machine unlearning in generative AI: A survey. \emph{arXiv preprint arXiv:2408.10862}.

\bibitem{schulman2017proximal}
Schulman, J., Wolski, F., Dhariwal, P., Radford, A., \& Klimov, O. (2017). Proximal policy optimization algorithms. \emph{arXiv preprint arXiv:1707.06347}.

\bibitem{haarnoja2018soft}
Haarnoja, T., Zhou, A., Abbeel, P., \& Levine, S. (2018). Soft actor-critic: Off-policy maximum entropy deep reinforcement learning with a stochastic actor. In \emph{International Conference on Machine Learning} (ICML) (pp. 1861-1870). PMLR.

\bibitem{heusel2017gans}
Heusel, M., Ramsauer, H., Unterthiner, T., Nessler, B., \& Hochreiter, S. (2017). GANs trained by a two time-scale update rule converge to a local Nash equilibrium. In \emph{Advances in Neural Information Processing Systems} (NeurIPS) (Vol. 30).

\bibitem{binkowski2018demystifying}
Bińkowski, M., Sutherland, D. J., Arbel, M., \& Gretton, A. (2018). Demystifying MMD GANs. In \emph{International Conference on Learning Representations} (ICLR).


\bibitem{wang2004image}
Wang, Z., Bovik, A. C., Sheikh, H. R., \& Simoncelli, E. P. (2004). Image quality assessment: From error visibility to structural similarity. \emph{IEEE Transactions on Image Processing}, 13(4), 600-612.

\end{thebibliography}
\end{document}